\documentclass[conference]{IEEEtran}
\IEEEoverridecommandlockouts
\usepackage[ruled,vlined]{algorithm2e}

\usepackage{cite}
\usepackage{amsmath,amssymb,amsfonts}
\usepackage{algorithmic}
\usepackage{graphicx}
\usepackage{textcomp}
\usepackage{xcolor}
\def\BibTeX{{\rm B\kern-.05em{\sc i\kern-.025em b}\kern-.08em
    T\kern-.1667em\lower.7ex\hbox{E}\kern-.125emX}}
    
\begin{document}


\title{SOFIM: Stochastic Optimization Using Regularized Fisher Information Matrix}

\author{Mrinmay Sen\IEEEauthorrefmark{1}\IEEEauthorrefmark{2}, A. K. Qin  \IEEEauthorrefmark{1}, \IEEEauthorblockN{Gayathri C\IEEEauthorrefmark{3}, Raghu Kishore N \IEEEauthorrefmark{3},  Yen-Wei Chen \IEEEauthorrefmark{4}, Balasubramanian Raman \IEEEauthorrefmark{5}}

\IEEEauthorblockA{\IEEEauthorrefmark{1}Dept. of Computing Technologies, Swinburne University of Technology, Hawthorn, Victoria, Australia}
\IEEEauthorblockA{\IEEEauthorrefmark{2}Dept. of Artificial Intelligence,  Indian Institute of Technology Hyderabad, Hyderabad, India}
\IEEEauthorblockA{\IEEEauthorrefmark{3}Dept. of Computer Science and Engineering, Mahindra University, Hyderabad, India}
\IEEEauthorblockA{\IEEEauthorrefmark{4}Dept. of Information Science and Engineering, Ritsumeikan University, Kyoto, Japan}
\IEEEauthorblockA{\IEEEauthorrefmark{5}Dept. of Computer Science and Engineering, Indian Institute of Technology Roorkee, Roorkee, India}

\IEEEauthorblockA{Emails: ai20resch11001@iith.ac.in, kqin@swin.edu.au,\\ \{se23ucse047, raghukishore.neelisetti\}@mahindrauniversity.edu.in, chen@is.ritsumei.ac.jp, bala@cs.iitr.ac.in}
}

\maketitle

\maketitle

\begin{abstract}
This paper introduces a new stochastic optimization method based on the regularized Fisher information matrix (FIM), named SOFIM, which can efficiently utilize the FIM to approximate the Hessian matrix for finding Newton’s gradient update in large-scale stochastic optimization of machine learning models. It can be viewed as a variant of natural gradient descent, where the challenge of storing and calculating the full FIM is addressed through making use of the regularized FIM and directly finding the gradient update direction via Sherman-Morrison matrix inversion. Additionally, like the popular Adam method, SOFIM uses the first moment of the gradient to address the issue of non-stationary objectives across mini-batches due to heterogeneous data. The utilization of the regularized FIM and Sherman-Morrison matrix inversion leads to the improved convergence rate with the same space and time complexities as stochastic gradient descent (SGD) with momentum. The extensive experiments on training deep learning models using several benchmark image classification datasets demonstrate that the proposed SOFIM outperforms SGD with momentum and several state-of-the-art Newton optimization methods in term of the convergence speed for achieving the pre-specified objectives of training and test losses as well as test accuracy.
\end{abstract}


\section{Introduction}
Stochastic optimization of probabilistic models are very much important in artificial intelligence (AI) applications like image classification, object detection, image segmentation etc. The optimization of probabilistic models is associated with optimization of a empirical probabilistic loss function $P(\mathbb{D}; \textbf{w})$ (i.e. negative log likelihood loss) defined as follows.
\begin{equation}
    \label{eq:1}
    \min_{\textbf{w}} P(\mathbb{D};\textbf{w})= \frac{1}{N}\sum_{i=1}^N p_i(\xi_i; \textbf{w})
\end{equation}
where, $\mathbb{D}= {\{\xi_i\}}$ is the set of data samples with N elements, $\textbf{w} \in {\mathbb{R}}^{d}$ is the set of model parameters to be estimated and $p_i(\xi_i; \textbf{w})$ is probabilistic loss (which is differentiable) for $i$-{th} data sample $\xi_i$. To solve the above problem, various iterative methods have been proposed, which can be grouped into two categories based on order of Taylor approximation \cite{taylor_Grosse_2021} of the loss function. First group is based on first-order stochastic gradient descent (SGD) \cite{sgd_Ketkar_2017} and another one is based on second-order Newton method \cite{newton_Jorge_1982}.

SGD is an iterative method of updating model parameters, where the gradient of a differentiable loss function  is used to find the update direction, which is shown in Eq. \ref{eq:2}.
\begin{equation}
    \label{eq:2}
    \textbf{w}_{t}= \textbf{w}_{t-1} - \eta_t \textbf{I}^{-1} \textbf{g}_t
\end{equation} 
where, subscript ``t" refers to $t$ - {th} iteration, $\textbf{I} \in \mathbb{R}^{d \times d}$ is an identity matrix, $\textbf{g}_t \in \mathbb{R}^{d \times 1}$ is the gradient of the loss function with respect to model parameters $\textbf{w}_{t-1}$ and $\eta_t$ is learning rate or step size, which is used for scaling the gradient. To improve the convergence of SGD in non-stationary settings, different modifications have been made, such that SGD with momentum \cite{sgd_with_momemntum}, AdaGrad \cite{adagrad}, SVRG \cite{svrg_Johnson}, Adam \cite{kingma2014adam} etc. Due to affordable linear time computational cost $O(Nd)$, SGD and its variants are well suited for large scale optimizations with huge data samples and large model. However, the major issue with these methods is the slow convergence rate. This is due to utilization of only gradient information, while updating model parameters. Also these methods are  highly sensitive to hyper-parameter choices. These limitations of first-order methods serve as motivation for us to employ the Newton method of optimization \cite{bottou2018optimization}, which offers a quadratic convergence rate and requires minimal hyper-parameter tuning \cite{polyak2007newton,second_order_review}.

In Newton method, the update direction is formulated by minimizing the second-order Taylor approximation of the loss function, which allows us to utilize Hessian curvature information along with gradient information while updating the model parameters as shown in Eq. \ref{eq:3}. 
\begin{equation}
    \label{eq:3}
    \textbf{w}_{t}= \textbf{w}_{t-1} - \eta {\textbf{H}_t}^{-1} \textbf{g}_t
\end{equation}
where, $\textbf{H}_t \in \mathbb{R}^{d \times d}$ is the Hessian of the loss function with respect to model parameters $\textbf{w}_{t-1}$. For large scale optimizations, calculation and storing of these Hessian \textbf{H} and its inverse $\textbf{H}^{-1}$ are the major challenges in Newton method. The computation of Hessian and its inverse are associated with a overall time complexity of $O(Nd^2 + d^3)$ and require a overall $O(d^2)$ space complexity, which may be impractical for large models and large datasets. To overcome these challenges of Newton method, researchers focus on utilization of approximated Hessian instead of the true Hessian. State-of-the-art Hessian approximation based Newton method of optimizations include Quasi-Newton method \cite{schoenberg2001optimization} and its variants (BFGS \cite{bonnans2006numerical}, L-BFGS \cite{liu1989limited}, oBFGS \cite{schraudolph2007stochastic}, SQN \cite{byrd2016stochastic}, SVRG-SQN \cite{moritz2016linearly}), AdaHessian \cite{yao2021adahessian}, NGD \cite{amari1998natural}, KFAC \cite{kfac_james},   Nyström-SGD \cite{singh2021nys} etc. BFGS approximates the inverse of the Hessian by using secant equation, which requires to store previous step's Hessian inverse. For storing of previous step's curvature information, BFGS needs $O(d^2)$ space complexity, which becomes the major bottleneck of BFGS for large scale application. As a solution of this issue of BFGS, L-BFGS comes into picture, which uses past difference of gradients and updates to approximate the Hessian inverse. L-BFGS can reduce the space complexity from $O(d^2)$ to O(md), where $1<m<<d$. The oBFGS algorithm is an extension of BFGS that employs stochastic gradients. SQN utilizes Hessian vector product while finding the Hessian approximation and SVRG-SQN is an extension of SQN, which uses variance-reduced gradients. Both SQN and SVRG-SQN may not be well suited for high-dimensional datasets due increased computational complexity. AdaHessian uses approximated Hessian diagonal instead of full Hessian. To approximate the Hessian diagonal, AdaHessian uses Hutchinson’s method. Nyström-SGD approximate the Hessian by using Nyström method on a partial column Hessian $\mathbb{C} \in \mathbb{R}^{d \times m}$, where $1<m<<d$ is randomly selected Hessian columns. Nyström-SGD does not require to store the full Hessian matrix, as it computes the update step directly using Sherman-Morrison formula of matrix inversion. Nyström-SGD has an overall time complexity of $O(mNd)$ and space complexity of $O(md)$, which are m times more than SGD. NGD is mostly similar to the Newton method, wherein the Fisher Information Matrix (FIM) is regarded as the equivalent of the Hessian for probabilistic loss functions. Same as Newton method, NGD has the issue of high time and space complexities for formulating and storing of the FIM, which arises a biggest question of applicability of NGD in large scale settings. To overcome these issues with NGD, KFAC comes into picture, where the individual FIM is calculated for each layer of the neural network and inversion of this FIM is made by using Kronecker product of two much smaller matrices. Even KFAC outperforms state-of-the-art NGD based algorithms, the computational costs are still greater than first-order based methods.  

This paper introduces SOFIM with the aims of efficiently utilizing Hessian curvature information in large-scale stochastic optimization of probabilistic models with the similar computational and space complexities as first-order methods. To achieve the same, SOFIM utilizes regularized Fisher information matrix as the Hessian of the loss function. As SOFIM uses Fisher information matrix for finding the Newton update, it can be viewed as a variant of natural gradient descent (NGD) \cite{amari1998natural}, where the problem of storing and calculation of full FIM are handled by using regularized FIM and directly finding the update direction using Sherman-Morrison formula of matrix inversion. Additionally, like Adam, SOFIM uses first moment of gradient to overcome the issue of non-stationary objectives across mini-batches, which is caused by heterogeneous data. Use of regularized FIM and Sherman-Morrison formula of matrix inversion lead to improved convergence rate in stochastic optimization in large scale settings with the same space and time complexities as SGD with momentum. Extensive experiments on several benchmark datasets with deep learning models show that SOFIM outperforms stochastic gradient descent (SGD) with momemtum and state-of-the-art Newton method of optimization methods such that Nystr\"om-SGD, L-BFGS and AdaHessian in term of faster convergence while achieving a certain precision of training $\&$ test losses and test accuracy.
The main contributions of SOFIM are as follows-
\begin{itemize}
  \item SOFIM uses regularized FIM as the Hessian of probabilistic loss function
  \item SOFIM uses Sherman-Morrison formula of matrix inversion to directly find the Newton update direction 
  \item Additionally, like Adam, SOFIM uses first moment of gradient to overcome the issue of non-stationary objectives across mini-batches which is caused by heterogeneous data 
\end{itemize}

\section{Preliminaries}
\subsection{Fisher information matrix (FIM)}

The Fisher information matrix \cite{ly2017tutorial} \textbf{F} $\in R^{d \times d}$ of a probabilistic model $P$, which maps to a conditional probability, is defined as 
\begin{equation}
    \label{eq:4}
    \textbf{F} = \underset{\{\xi_i\}}{\mathbb{E}}[\nabla \log p(\xi_i;\textbf{w}) \nabla \log p(\xi_i; \textbf{w})^T]\\
    =\mathbb{E}[\textbf{g}^i {\textbf{g}^i}^T]
\end{equation}
where, $\mathbb{D}=\{\xi_i\}$ is observable random variables, $\textbf{w}$ is the set of unknown  parameters to be estimated and  $\textbf{g}^i$ = $\nabla \log p(\xi_i; \textbf{w})$ is the gradient of logarithm of the probabilistic loss function (i.e. gradient of log likelihood) for data sample $\xi_i$. It can be proved that, for probabilistic models, the  FIM (\textbf{F}) can be treated as the negative expected Hessian \cite{martens2020new}.

\subsection{Natural gradient descent (NGD)}
NGD is similar to Newton method of optimization, where the FIM of a negative log probabilistic model i.e. negative log likelihood is considered as Hessian. NGD update rule is derived by replacing \textbf{H} with  \textbf{F} in Eq. \ref{eq:3}, as shown in Eq. \ref{eq:5}. 

\begin{equation}
    \label{eq:5}
    \textbf{w}_{t}= \textbf{w}_{t-1} - \eta \textbf{F}_t^{-1} \textbf{g}_t
\end{equation}

The main challenges with NGD are associated with the computation and storing of \textbf{F} and its inverse that need cubic computational complexity and quadratic space complexity.

\subsection{Sherman Morrison formula of matrix inversion}
Let \textbf{A} $\in \mathbb{R}^{d \times d}$ be a invertible square matrix and \textbf{u}, \textbf{v} $\in \mathbb{R}^{d \times 1}$ are column vectors. Then, by using Sherman Morrison formula of matrix inversion \cite{sherman1950adjustment}, we can directly compute the inverse of the the matrix (\textbf{A} + $\textbf{uv}^T$) $\in \mathbb{R}^{d \times d}$ as follows. 

\begin{equation}
    \label{eq:6}
    (\textbf{A} + \textbf{uv}^T)^{-1}= \textbf{A}^{-1} - \frac{\textbf{A}^{-1}\textbf{uv}^T \textbf{A}^{-1}}{1+\textbf{v}^T \textbf{A}^{-1}\textbf{u}}
\end{equation}

\begin{algorithm}
\SetKwData{Left}{left}
\SetKwData{This}{this}
\SetKwData{Up}{up}
\SetKwFunction{Union}{Union}
\SetKwFunction{FindCompress}{FindCompress}
\SetKwInOut{Input}{input}
\SetKwInOut{Output}{output}
\caption{SOFIM}
\Input{$T$: Iterations, $\textbf{w}_0$: Randomly initialized model, $\eta$: learning rate, $\rho$: FIM regularization parameter, $\beta \in [0, 1)$: Exponential decay rates for the moment estimate, $M_0 \leftarrow 0$ :  Initial moment vector, which is initialized with zero}
\Output{Updated model $\textbf{w}_{t}$}
\For{$t\leftarrow 1$ \KwTo $T$}{
\emph{Finds stochastic gradient $\textbf{g}_t \in \mathbb{R}^{d \times 1}$ \\ Finds first moment of $\textbf{g}_t$ as $M_t$= $\beta M_{t-1} + (1-\beta) \textbf{g}_t$\\ Do bias correction $\widehat{M}_t= \frac{M_t}{1-\beta^t}$, here $\beta^t$ is $\beta$ to the power t\\Finds ${\textbf{F}_t}^{-1} \widehat{M}_t$=$\frac{\widehat{M}_t}{\rho}$ - $\frac{\frac{\widehat{M}_t {\widehat{M}_t}^T \widehat{M}_t}{\rho^2}}{1+\frac{{\widehat{M}_t}^T \widehat{M}_t}{\rho}}$, where $\textbf{F}_t = \widehat{M}_t \widehat{M}_t^T + \rho \textbf{I}$ \\Finds updated model $\textbf{w}_t=\textbf{w}_{t-1}-\eta {\textbf{F}_t}^{-1} \widehat{M}_t$}
}
\label{alg:algo1}
\end{algorithm}

\section{Proposed method}
One iteration of SOFIM is shown in Algo. \ref{alg:algo1}. In each iteration, SOFIM first finds stochastic gradient $\textbf{g}_t=\mathbb{E}[\nabla (-\log p_i(\xi_i; \textbf{w}_{t-1}))]$, where $-\log p_i(\xi_i; \textbf{w}_{t-1})$ is negative logarithm of probabilistic loss $p_i (\xi_i; \textbf{w}_{t-1})$ or negative log likelihood (where $\xi_i \in \mathbb{D}^t$ and $\mathbb{D}^t \subseteq \mathbb{D}$). Then, SOFIM utilizes regularized FIM with first moment $M_t$ (with exponential decay rate $\beta$ $\in$ [0, 1)) of this stochastic gradient $\textbf{g}_t$ to find the Newton update. Like Adam, SOFIM uses bias corrected estimate $\widehat{M}_t$ of the first moment $M_t$. 

\subsection{Regularized fisher information matrix (FIM)}
The empirical FIM of a negative log probabilistic function is defined as \textbf{F}=$\mathbb{E} [\textbf{g}^i {\textbf{g}^i}^T]$, which is equivalent to expected Hessian. For large scale settings, it may be impractical to calculate this \textbf{F} due to requirements of large time complexity of $O(Nd^3)$ and space complexity of $O(d^2)$. To overcome this challenge, SOFIM utilizes the following regularized variant of empirical FIM  as shown in Eq. \ref{eq:7} 

\begin{equation}
    \label{eq:7}
    \textbf{F} = \mathbb{E} [\textbf{g}^i {\textbf{g}^i}^T] \equiv \mathbb{E}[\textbf{g}^i] {\mathbb{E}[{\textbf{g}^i}]}^T + \rho \textbf{I}
\end{equation}
where, $\rho $ is regularization term and \textbf{I} $\in R^{d \times d}$ is Identity matrix.

\subsection{Calculate Newton update using regularized FIM}
Once the stochastic gradient $\textbf{g}_t=\mathbb{E} [\textbf{g}^i]$ is calculated, SOFIM finds the first moment of $\textbf{g}_t$ as $M_t = \beta M_{t-1} + (1-\beta) \textbf{g}_t$ and its bias corrected estimate $\widehat{M}_t$, which are inspired from the paper of Adam. This bias corrected estimate of gradient is now used for finding the Newton update. SOFIM uses this $\widehat{M}_t$ to find the FIM as $\textbf{F}_t = \widehat{M}_t {\widehat{M}_t}^T + \rho \textbf{I}$ and uses this $\textbf{F}_t$ in Eq. \ref{eq:5} to update the model parameters as shown in Eq. \ref{eq:8}.
\begin{equation}
    \label{eq:8}
    \textbf{w}_{t} = \textbf{w}_{t-1} - \eta {(\widehat{M_t} {\widehat{M_t}^T} + \rho \textbf{I})}^{-1} \widehat{M_t}
\end{equation}

To directly find the update direction ${\textbf{F}_t}^{-1} \widehat{M_t}$, SOFIM uses Eq. \ref{eq:6} of Sherman Morrison formula of matrix inversion. SOFIM replaces \textbf{A} with $\rho \textbf{I}$ and \textbf{u} $\&$ \textbf{v} with $\widehat{M_t}$ in Eq. \ref{eq:6} and finds the update direction as shown in Eq. \ref{eq:9}.

\begin{equation}
    \label{eq:9}
    {\textbf{F}_t}^{-1} \widehat{M_t} = \frac{\widehat{M}_t}{\rho} - \frac{\frac{\widehat{M}_t {\widehat{M}_t}^T \widehat{M}_t}{\rho^2}}{1+\frac{{\widehat{M}_t}^T \widehat{M}_t}{\rho}}
\end{equation}

\begin{figure*}[!]
  \centering
  \includegraphics[width=1\linewidth]{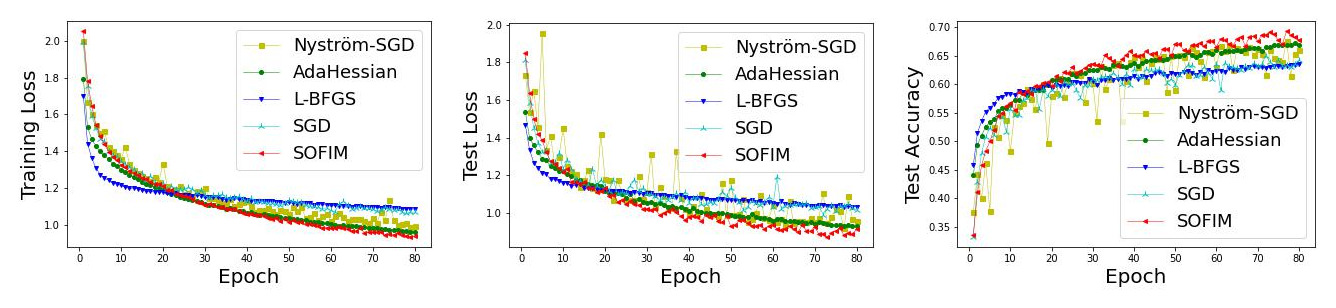}
  \caption{Iteration-wise comparisons of various methods regarding training loss, test loss, and test accuracy on CIFAR10 using the LeNet5 model.}\label{fig:p1}
\end{figure*}
\begin{figure*}[!]
  \centering
  \includegraphics[width=1\linewidth]{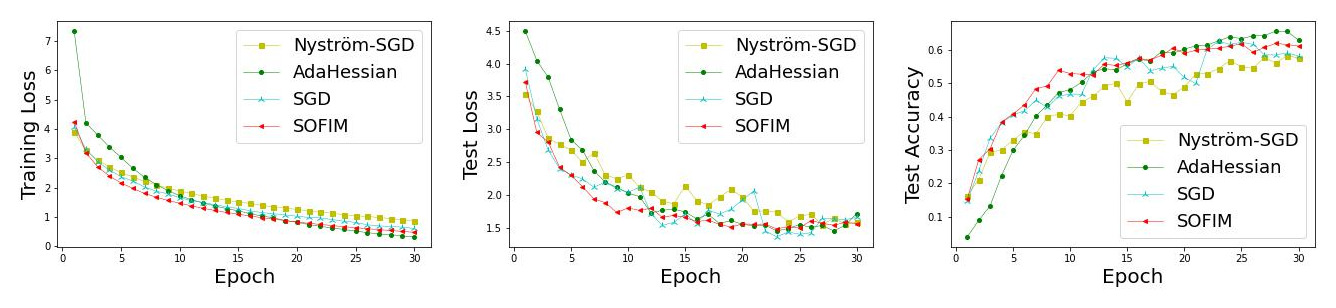}
  \caption{Iteration-wise comparisons of various methods regarding training loss, test loss, and test accuracy on CIFAR100 using the Resnet9 model.}\label{fig:p2}
\end{figure*}

\begin{figure*}[!]
  \centering
  \includegraphics[width=1\linewidth]{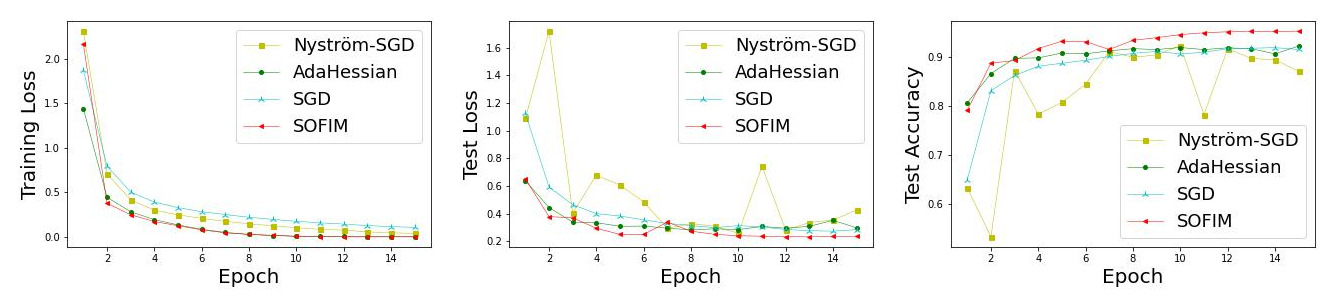}
  \caption{Iteration-wise comparisons of various methods regarding training loss, test loss, and test accuracy on SVHN using the Resnet9 model}\label{fig:p3}
\end{figure*}

\begin{figure*}[!]
  \centering
  \includegraphics[width=1\linewidth]{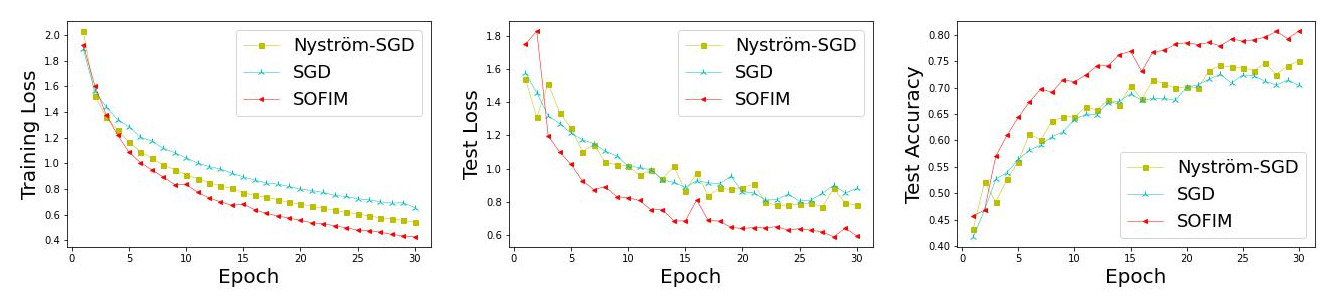}
  \caption{Iteration-wise comparisons of various methods regarding training loss, test loss, and test accuracy on CIFAR10 using the  Resnet18 model}\label{fig:p4}
\end{figure*}

\begin{figure*}[!]
  \centering
  \includegraphics[width=1\linewidth]{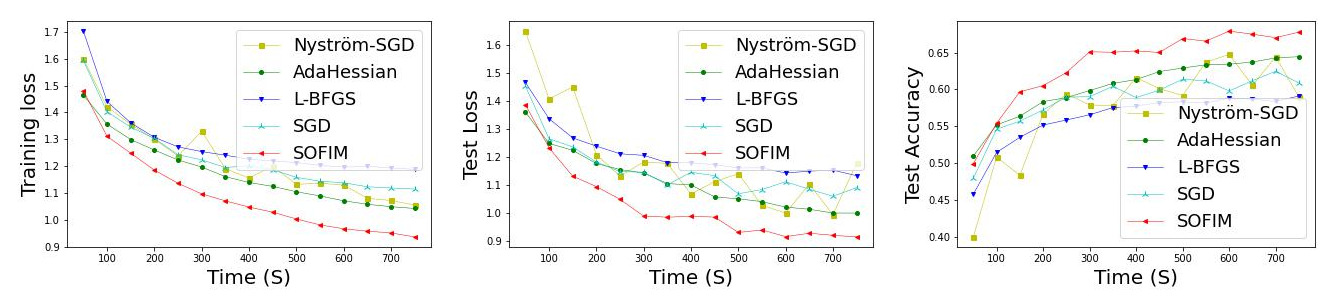}
  \caption{Time comparisons of various methods regarding training loss, test loss, and test accuracy on CIFAR10 using the LeNet5 model.}\label{fig:p5}
\end{figure*}

\begin{figure*}[!]
  \centering
  \includegraphics[width=1\linewidth]{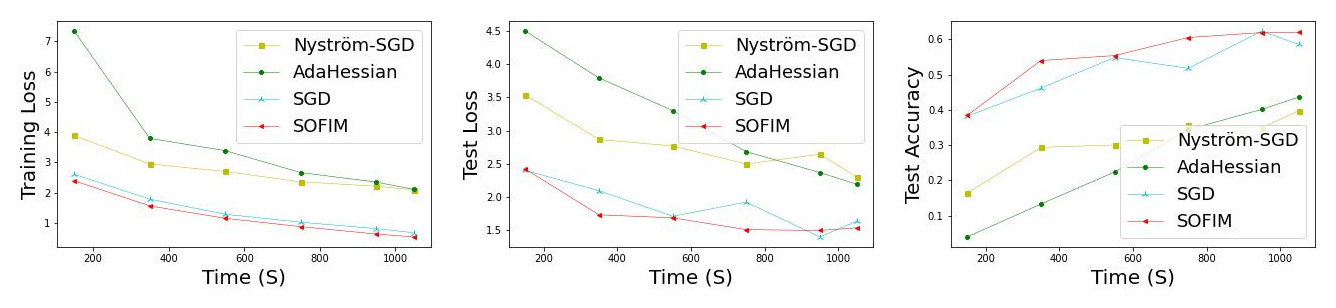}
  \caption{Time comparisons of various methods regarding training loss, test loss, and test accuracy on CIFAR100 using the Resnet9 model.}\label{fig:p6}
\end{figure*}

\begin{figure*}[!]
  \centering
  \includegraphics[width=1\linewidth]{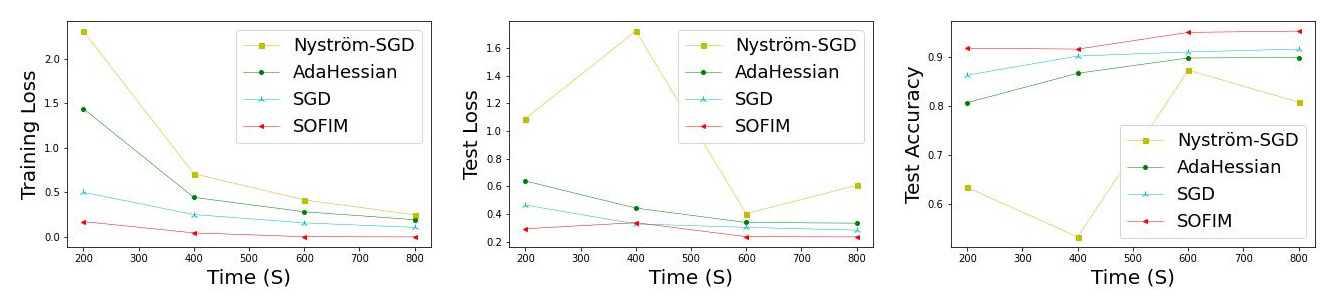}
  \caption{Time comparisons of various methods regarding training loss, test loss, and test accuracy on SVHN using the Resnet9 model}\label{fig:p7}
\end{figure*}

\begin{figure*}[!]
  \centering
  \includegraphics[width=1\linewidth]{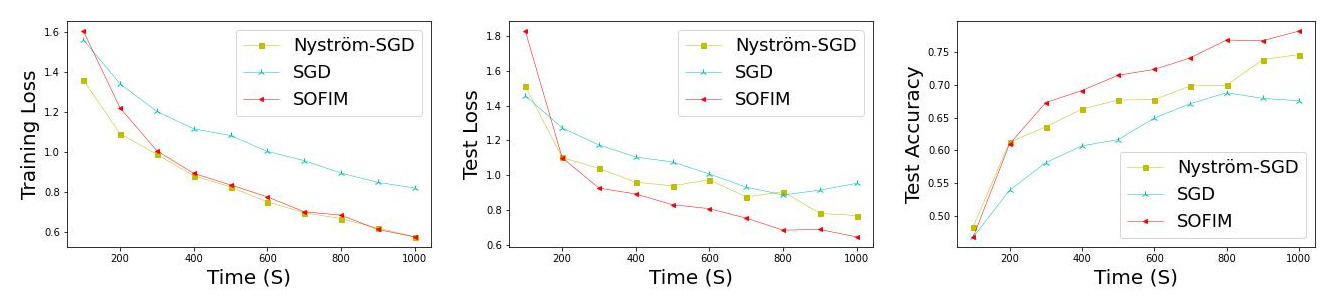}
  \caption{Time comparisons of various methods regarding training loss, test loss, and test accuracy on CIFAR10 using the  Resnet18 model}\label{fig:p8}
\end{figure*}
\begin{figure*}[!]
  \centering
  \includegraphics[width=1\linewidth]{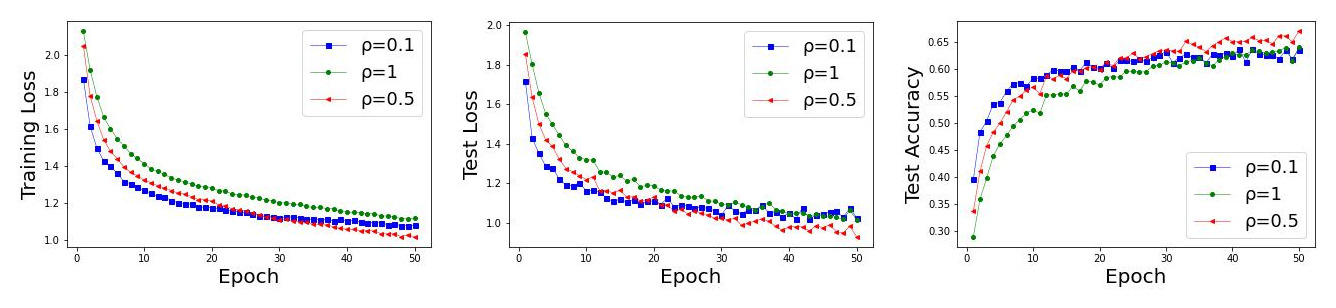}
  \caption{Effect of $\rho$ on training with SOFIM of LeNet5 model on CIFAR10.}\label{fig:p9}
\end{figure*}
\subsection{Complexities}
In SOFIM, calculation of stochastic gradient and its moment's estimate need $O(2d)$ space complexities and overall $O(Nd)$ time complexity. Finding Newton update with Sherman Morrison formula of matrix inversion requires both $O(d)$ time and space complexities. So, we can conclude that the overall time complexity of SOFIM is $O(d)$ and the overall space complexity is $O(2d)$, which are same as SGD with momentum.

\subsection{Convergence guarantee for convex loss function}
From the Theorem 4 of the paper of Haishan et al. \cite{ye2017approximate}, we can claim that for convex loss function, SOFIM can enjoy a linear-quadratic convergence rate, as SOFIM uses a regularized variant of Newton method of optimization with the approximated Hessian is in form of $(\textbf{A} + \textbf{uu}^T)$, where, \textbf{A}= $\rho \textbf{I}$ is a positive semi-definite matrix and $\textbf{u}=\widehat{M}_t \in \mathbb{R}^{d \times 1}$. 

\section{Experimental setup}
To validate the performance of SOFIM, , we conduct extensive experiments on image classification tasks of CIFAR10, CIFAR100 and SVHN datasets. For CIFAR10 classification, we use LeNet5 and Resnet18 deep learning models. For the classification tasks of CIFAR100 and SVHN datasets, we use Resnet9 deep learning model. We use cross entropy loss function (which is a negative log likelihood loss) for these classification tasks. We compare our algorithm with SGD with momentum and state-of-the-art Newton method of optimization algorithms such that Nystr\"om-SGD, L-BFGS $\&$ AdaHessian. We conduct our experiments on different sets of hyper-parameters for each method and find the best performing model by considering minimum training $\&$ test losses and maximum test accuracy. We use learning rate or step size $\eta \in \{1, 0.1, 0.01, 0.001, 0.0001\}$ for all the methods. For SGD we use momentum=0.9, weight decay = $1e-6$ and pytorch CosineAnnealingLR scheduler of learning rate. For Nystr\"om-SGD, we use number of selected Hessian column $m \in \{5, 10\}$ for Resnet18 $\&$ Resnet9 models and m=30 for LeNet5 model, Hessian update frequency $l=3$, Hessian regularization parameter $\rho \in \{1, 0.5, 0.1\}$. For L-BFGS, we use maximum iterations $\in \{4, 5\}$. For AdaHessian, we use $\beta_1 = 0.9$, $\beta_2 = 0.99$ and Hessian power=1. For, SOFIM, we use gradient moment's parameter $\beta = 0.9$, FIM regularization parameter $\rho \in \{1, 0.5, 0.1\}$. For all the methods, we use a mini-batch size = 512. We implement all the methods using Tesla V100 GPU and PyTorch-1.12.1+cu102. We use same settings and same initialization for all the methods, while doing comparisons.

\subsection{Results}
Our experimental results and comparisons have been depicted in figs. \ref{fig:p1}, \ref{fig:p2}, \ref{fig:p3}, \ref{fig:p4}, \ref{fig:p5}, \ref{fig:p6}, \ref{fig:p7} and \ref{fig:p8}. From these figures, it may be observed that, in SOFIM, the training $\&$ test losses are getting decreased faster (in terms of both the time and iterations) than SGD and state-of-the-art Newton methods such that Nystr\"om-SGD, L-BFGS $\&$ AdaHessian. From these figures, It may also be observed that the test accuracy achieved by SOFIM are better than Existing methods. As we use same settings and same initialization for all the methods, while doing comparisons, we may cliam that SOFIM can outperform SGD, Nystr\"om-SGD, L-BFGS and AdaHessian in term of faster convergence while achieving a certain precision of training $\&$ test losses and test accuracy. From our experiments, we noticed that Nystr\"om-SGD  can not able to perform well for LeNet5 and Resnet9 models. The same thing, we noticed for L-BFGS also. The reason may be that Nyström-SGD and L-BFGS require a greater number of selected Hessian columns m and ${iters}_{max}$ respectively, which results in incresed time and space complexities as compared to SOFIM.

\subsection{Effect of $\rho$}
The effect of $\rho$ on the performance of SOFIM has been analysed in fig. \ref{fig:p9}. This analysis is made on image classification tasks of CIFAR10 dataset with LeNet5 model. From this fig.\ref{fig:p9}, it may be observed that, for $\rho = 0.5 $, SOFIM performs well. Even from the experiments on image classification tasks of  CIFAR100 with Resnet9, SVHN with Resnet9 and CIFAR10 with Resnet18 , we also noticed that for $\rho =0.5$, SOFIM can perform well. From our experiments, we observed that, for $\rho < 0$, SOFIM performs very poor.  

\section{Conclusions and Future work}
\label{sec:conclusion}
We proposed a new stochastic optimization method named SOFIM, aiming to accelerate the convergence speed of training a probabilistic model while keeping the space and time complexities as those in SGD with momentum. SOFIM features the regularized FIM used to approximate the Hessian matrix and the Sherman-Morrison formulation of matrix inversion used to directly compute the gradient update direction. Also, SOFIM uses the first moment of the gradient, like Adam, to handle the issue of non-stationary objectives across mini-batches, caused by heterogeneous data. Experimental results have validated the faster convergence speed of SOFIM compared to SGD with momentum and several state-of-the-art Newton optimization methods, such as Nystr\"om-SGD, L-BFGS and AdaHessian. In the future, we plan to apply SOFIM to solve versatile machine learning tasks, e.g., \cite{qin2005initialization}, to hybridize it with popular evolutionary algorithms \cite{qin2013differential} to better address evolutionary learning tasks, e.g., \cite{gong2016discrete}, and to comprehensively evaluate its efficacy and meanwhile identify its shortcomings for further improvement.

\section*{Acknowledgment}

This work is supported in part by Grant-in-Aid for Scientific Research from the Japanese Ministry for Education, Science, Culture, and Sports (MEXT) under Grant No. 20KK0234 (co-author: Yen-Wei Chen), and the Australian Research Council (ARC) under Grant No. LP180100114 and DP200102611.

\bibliographystyle{IEEEtran}
{\small
\bibliography{ref}}

\begin{thebibliography}{10}
\providecommand{\url}[1]{#1}
\csname url@samestyle\endcsname
\providecommand{\newblock}{\relax}
\providecommand{\bibinfo}[2]{#2}
\providecommand{\BIBentrySTDinterwordspacing}{\spaceskip=0pt\relax}
\providecommand{\BIBentryALTinterwordstretchfactor}{4}
\providecommand{\BIBentryALTinterwordspacing}{\spaceskip=\fontdimen2\font plus
\BIBentryALTinterwordstretchfactor\fontdimen3\font minus \fontdimen4\font\relax}
\providecommand{\BIBforeignlanguage}[2]{{%
\expandafter\ifx\csname l@#1\endcsname\relax
\typeout{** WARNING: IEEEtran.bst: No hyphenation pattern has been}%
\typeout{** loaded for the language `#1'. Using the pattern for}%
\typeout{** the default language instead.}%
\else
\language=\csname l@#1\endcsname
\fi
#2}}
\providecommand{\BIBdecl}{\relax}
\BIBdecl

\bibitem{taylor_Grosse_2021}
R.~Grosse, ``Taylor approximations,'' \emph{Neural Network Training Dynamics. Lecture Notes, University of Toronto}, 2021.

\bibitem{sgd_Ketkar_2017}
N.~Ketkar and N.~Ketkar, ``Stochastic gradient descent,'' \emph{Deep learning with Python: A hands-on introduction}, pp. 113--132, 2017.

\bibitem{newton_Jorge_1982}
J.~J. Mor{\'e} and D.~C. Sorensen, ``Newton's method,'' Argonne National Lab., IL (USA), Tech. Rep., 1982.

\bibitem{sgd_with_momemntum}
A.~Cutkosky and H.~Mehta, ``Momentum improves normalized sgd,'' in \emph{International conference on machine learning}.\hskip 1em plus 0.5em minus 0.4em\relax PMLR, 2020, pp. 2260--2268.

\bibitem{adagrad}
J.~Duchi, E.~Hazan, and Y.~Singer, ``Adaptive subgradient methods for online learning and stochastic optimization.'' \emph{Journal of machine learning research}, vol.~12, no.~7, 2011.

\bibitem{svrg_Johnson}
R.~Johnson and T.~Zhang, ``Accelerating stochastic gradient descent using predictive variance reduction,'' \emph{Advances in neural information processing systems}, vol.~26, 2013.

\bibitem{kingma2014adam}
D.~P. Kingma and J.~Ba, ``Adam: A method for stochastic optimization,'' \emph{arXiv preprint arXiv:1412.6980}, 2014.

\bibitem{bottou2018optimization}
L.~Bottou, F.~E. Curtis, and J.~Nocedal, ``Optimization methods for large-scale machine learning,'' \emph{SIAM review}, vol.~60, no.~2, pp. 223--311, 2018.

\bibitem{polyak2007newton}
B.~T. Polyak, ``Newton’s method and its use in optimization,'' \emph{European Journal of Operational Research}, vol. 181, no.~3, pp. 1086--1096, 2007.

\bibitem{second_order_review}
H.~H. Tan and K.~H. Lim, ``Review of second-order optimization techniques in artificial neural networks backpropagation,'' in \emph{IOP conference series: materials science and engineering}, vol. 495, no.~1.\hskip 1em plus 0.5em minus 0.4em\relax IOP Publishing, 2019, p. 012003.

\bibitem{schoenberg2001optimization}
R.~Schoenberg, ``Optimization with the quasi-newton method,'' \emph{Aptech Systems Maple Valley WA}, pp. 1--9, 2001.

\bibitem{bonnans2006numerical}
J.-F. Bonnans, J.~C. Gilbert, C.~Lemar{\'e}chal, and C.~A. Sagastiz{\'a}bal, \emph{Numerical optimization: theoretical and practical aspects}.\hskip 1em plus 0.5em minus 0.4em\relax Springer Science \& Business Media, 2006.

\bibitem{liu1989limited}
D.~C. Liu and J.~Nocedal, ``On the limited memory bfgs method for large scale optimization,'' \emph{Mathematical programming}, vol.~45, no. 1-3, pp. 503--528, 1989.

\bibitem{schraudolph2007stochastic}
N.~N. Schraudolph, J.~Yu, and S.~G{\"u}nter, ``A stochastic quasi-newton method for online convex optimization,'' in \emph{Artificial intelligence and statistics}.\hskip 1em plus 0.5em minus 0.4em\relax PMLR, 2007, pp. 436--443.

\bibitem{byrd2016stochastic}
R.~H. Byrd, S.~L. Hansen, J.~Nocedal, and Y.~Singer, ``A stochastic quasi-newton method for large-scale optimization,'' \emph{SIAM Journal on Optimization}, vol.~26, no.~2, pp. 1008--1031, 2016.

\bibitem{moritz2016linearly}
P.~Moritz, R.~Nishihara, and M.~Jordan, ``A linearly-convergent stochastic l-bfgs algorithm,'' in \emph{Artificial Intelligence and Statistics}.\hskip 1em plus 0.5em minus 0.4em\relax PMLR, 2016, pp. 249--258.

\bibitem{yao2021adahessian}
Z.~Yao, A.~Gholami, S.~Shen, M.~Mustafa, K.~Keutzer, and M.~Mahoney, ``Adahessian: An adaptive second order optimizer for machine learning,'' in \emph{proceedings of the AAAI conference on artificial intelligence}, vol.~35, no.~12, 2021, pp. 10\,665--10\,673.

\bibitem{amari1998natural}
S.-I. Amari, ``Natural gradient works efficiently in learning,'' \emph{Neural computation}, vol.~10, no.~2, pp. 251--276, 1998.

\bibitem{kfac_james}
J.~Martens and R.~Grosse, ``Optimizing neural networks with kronecker-factored approximate curvature,'' in \emph{International conference on machine learning}.\hskip 1em plus 0.5em minus 0.4em\relax PMLR, 2015, pp. 2408--2417.

\bibitem{singh2021nys}
D.~Singh, H.~Tankaria, and M.~Yamada, ``Nys-newton: Nystr$\backslash$" om-approximated curvature for stochastic optimization,'' \emph{arXiv preprint arXiv:2110.08577}, 2021.

\bibitem{ly2017tutorial}
A.~Ly, M.~Marsman, J.~Verhagen, R.~P. Grasman, and E.-J. Wagenmakers, ``A tutorial on fisher information,'' \emph{Journal of Mathematical Psychology}, vol.~80, pp. 40--55, 2017.

\bibitem{martens2020new}
J.~Martens, ``New insights and perspectives on the natural gradient method,'' \emph{The Journal of Machine Learning Research}, vol.~21, no.~1, pp. 5776--5851, 2020.

\bibitem{sherman1950adjustment}
J.~Sherman and W.~J. Morrison, ``Adjustment of an inverse matrix corresponding to a change in one element of a given matrix,'' \emph{The Annals of Mathematical Statistics}, vol.~21, no.~1, pp. 124--127, 1950.

\bibitem{ye2017approximate}
H.~Ye, L.~Luo, and Z.~Zhang, ``Approximate newton methods and their local convergence,'' in \emph{International Conference on Machine Learning}.\hskip 1em plus 0.5em minus 0.4em\relax PMLR, 2017, pp. 3931--3939.

\bibitem{qin2005initialization}
A.~K. Qin and P.~N. Suganthan, ``Initialization insensitive {LVQ} algorithm based on cost-function adaptation,'' \emph{Pattern Recognition}, vol.~38, no.~5, pp. 773--776, 2005.

\bibitem{qin2013differential}
A.~K. Qin and X.~Li, ``Differential evolution on the {CEC-2013} single-objective continuous optimization testbed,'' in \emph{2013 IEEE Congress on Evolutionary Computation}, 2013, pp. 1099--1106.

\bibitem{gong2016discrete}
M.~Gong, Y.~Wu, Q.~Cai, W.~Ma, A.~K. Qin, Z.~Wang, and L.~Jiao, ``Discrete particle swarm optimization for high-order graph matching,'' \emph{Information Sciences}, vol. 328, pp. 158--171, 2016.

\end{thebibliography}
\end{document}